\documentclass[journal]{IEEEtran}
\usepackage{cite}

\ifCLASSINFOpdf
\else
\fi
\usepackage[cmex10]{amsmath}
\usepackage{algorithmic}

\usepackage{array}

\usepackage[tight,footnotesize]{subfigure}
\usepackage{url}
\usepackage{amssymb}

\usepackage{siunitx}
\usepackage{graphicx}

\usepackage[normalem]{ulem}
\usepackage{xcolor}
\usepackage{color}
\usepackage{blindtext}
\newcommand{\rpm}{\raisebox{.2ex}{$\scriptstyle\pm$}}

\begin{document}
%

\title{A Mosquito Pick-and-Place System for PfSPZ-based Malaria Vaccine Production}

%
%
%

\author{Henry~Phalen*,
        Prasad~Vagdargi*,
        Mariah~L.~Schrum,
        Sumana~Chakravarty,
        Amanda~Canezin,
        Michael~Pozin,
        Suat~Coemert,
        Iulian~Iordachita,~\IEEEmembership{Senior Member, IEEE},
        Stephen~L.~Hoffman,       
        Gregory~S.~Chirikjian,~\IEEEmembership{Fellow, IEEE},
        Russell~H.~Taylor,~\IEEEmembership{Life Fellow, IEEE}
\thanks{*The first two authors contributed equally to this work.}
\thanks{H. Phalen, P. Vagdargi, G. S. Chirikjian, I. Iordachita, and R. H. Taylor are with the Laboratory for Computational Sensing and Robotics (LCSR) at the Johns Hopkins University in Baltimore, MD, USA. Previously, M. L. Schrum, A. Canezin, M. Pozin, and S. Coemert were also with the LCSR. Now, M. L. Schrum is with Georgia Tech, A. Canezin is with Accenture, M. Pozin is with Auris Health, and S. Coemert is with the Technical University of Munich. G. S. Chirikjian is currently at the National University of Singapore, Department of Mechanical Engineering, 117575, Singapore. S. Chakravarty and S. L. Hoffman are with Sanaria, Inc. in Rockville, MD, USA.}
\thanks{A supporting video file has been provided with this manuscript.}}

%
%


\markboth{Manuscript for Special Issue of IEEE CASE 2019 for IEEE T-ASE}%
{Shell \MakeLowercase{\textit{et al.}}: Bare Demo of IEEEtran.cls for Journals}
%



\maketitle

\begin{abstract}
The treatment of malaria is a global health challenge that stands to benefit from the widespread introduction of a vaccine for the disease. A method has been developed to create a live organism vaccine using the sporozoites (SPZ) of the parasite \textit{Plasmodium falciparum} (Pf), which are concentrated in the salivary glands of infected mosquitoes. Current manual dissection methods to obtain these PfSPZ are not optimally efficient for large-scale vaccine production. We propose an improved dissection procedure and a mechanical fixture that increases the rate of mosquito dissection and helps to deskill this stage of the production process. We further demonstrate the automation of a key step in this production process, the picking and placing of mosquitoes from a staging apparatus into a dissection assembly. This unit test of a robotic mosquito pick-and-place system is performed using a custom-designed micro-gripper attached to a four degree of freedom (4-DOF) robot under the guidance of a computer vision system. Mosquitoes are autonomously grasped and pulled to a pair of notched dissection blades to remove the head of the mosquito, allowing access to the salivary glands. Placement into these blades is adapted based on output from computer vision to accommodate for the unique anatomy and orientation of each grasped mosquito. In this pilot test of the system on 50 mosquitoes, we demonstrate a 100\% grasping accuracy and a 90\% accuracy in placing the mosquito with its neck within the blade notches such that the head can be removed. This is a promising result for this difficult and non-standard pick-and-place task.\end{abstract}

Note to Practitioners:
\begin{abstract}
Automated processes could help increase malaria vaccine production to global scale. Currently, production requires technicians to manually dissect mosquitoes, a process that is slow, tedious, and requires a lengthy training regimen. This paper presents an an improved manual fixture and procedure that reduces technician training time. Further, an approach to automate this dissection process is proposed and the critical step of robotic manipulation of the mosquito with the aid of computer vision is demonstrated. Our approach may serve as a useful example of system design and integration for practitioners that seek to perform new and challenging pick-and-place tasks with small, non-uniform, and highly deformable objects.
\end{abstract}

Primary and Secondary Keywords
\begin{IEEEkeywords}
Biomedical engineeering, Robots, Robot vision systems, Manufacturing automation, Biomedical imaging
\end{IEEEkeywords}

%
\IEEEpeerreviewmaketitle

\section{Introduction}
%
%
%
%

    \IEEEPARstart{M}{alaria} presents a tremendous public health burden. The World Health Organization estimates 228 million individuals worldwide were infected with malaria in 2018, and with an estimated 405,000 deaths, the disease is among the top 20 leading causes of death globally among both adults and infants \cite{WHOreport2019, owidcausesofdeath}. With increasing drug and insecticide resistance, it has become difficult for current treatments to maintain efficacy in reducing the prevalence of malaria worldwide \cite{WHOstrat}. Development of malarial vaccines present a promising way forward in the global effort for malaria eradication \cite{WHOstrat}.
    
    Humans become infected with malaria-causing parasites when \textit{Anopheles} mosquitoes inoculate the sporozoite (SPZ) developmental stage of the parasite. SPZ reside in mosquito salivary glands immediately prior to transmission during feeding.  Progress has been made in the development of the Sanaria \textit{Plasmodium falciparum} (Pf) sporozoite-based vaccine (Sanaria\textsuperscript{\textregistered} PfSPZ Vaccine), an effective vaccine manufactured from PfSPZ extracted from the salivary glands of female \textit{Anopheles} mosquitoes \cite{meta,ishizuka,epstein,sissoko2017safety,lyke2017attenuated, mordmuller2017sterile}. Such a vaccine may reduce the burden of the disease by providing immunity against Pf, the most common malarial parasite, which was estimated to account for greater than 95\% of deaths caused by malaria in 2017 \cite{WHOreport2018}. These vaccines have shown a high level of protective efficacy against controlled human malaria infection and malaria transmitted in the field, making them ideal for large-scale malaria elimination campaigns in geographically defined malarious regions. However, a limiting step in the manufacture of PfSPZ-based vaccines has been the extraction of the salivary glands and isolation of sporozoites from very large numbers of infected mosquitoes to meet expected demand for the vaccine. 
    
    The process of salivary gland dissection has only been demonstrated with training-intensive manual processes. In these traditional methods, technicians are presented with freshly-sacrificed, lab-grown mosquitoes and process them under a microscope, one at a time. To gain access to the glands, a technician first removes the mosquito head using the beveled edge of a hypodermic needle as a knife. Next, the technician gently squeezes the mosquito body to remove from the thorax a volume of exudate that includes the PfSPZ-laden salivary glands. The exudate from mosquitoes is collected and processed for the isolation of PfSPZ. Technicians require extensive training to be proficient at dissection and have a  throughput of around 4-5 mosquitoes per minute on average. In order be certified for manual dissection at Sanaria, untrained personnel go through a rigorous training procedure involving 1-3 one-hour sessions every week. Each trainee is graded on metrics including cleanliness of the exudate, throughput, and SPZ yields from infected mosquitoes. Training is considered complete if the trainee’s throughput is approximately 250 mosquitoes dissected per hour (Mdph) and the sporozoite yields are within 25\% of a certified dissector. The time to complete this training has varied tremendously between operators depending on dexterity and hand-eye coordination skills for successful micromanipulation of mosquitoes under a stereomicroscope. Although further entrainment does occur for every operator as they continue participation in dissection, the gestation period prior to qualification is too variable and long, averaging around 16 weeks.    

    The automation of salivary gland harvesting from mosquitoes has been attempted, to our knowledge, only one time in the past. This included an effort to fully automate the salivary gland extraction production process by means of a robot. Although there is very little technical material publicly available about this system beyond a YouTube video included as part of a fundraising effort \cite{youtube} and news  articles \cite{Lap,bor}, the proposed system evidently included a computer vision subsystem to locate mosquitoes in a dish and a Cartesian (XYZ) robot to position an end effector to grasp the mosquito and feed each one sequentially into a tube for further processing. However, no literature supports the success of any such process at this time.

    Our group proposes a novel system, automated end-to-end, to perform the mosquito dissection for PfSPZ vaccine production. This is a complex system with several components, each targeting a stage of the dissection process. The main components include a feeder mechanism for sorting and presenting mosquito bodies, a dissector which automatically removes the mosquito head and squeezes and harvests the exudate, and a pick-and-place robot that manipulates mosquito bodies from the feeder to the dissector. Further, several computer vision implementations are used in the control and verification of these components' functions.

   To help define formal design requirements and clarify the design feasibility and priority for components within this larger system, we first designed \cite{patent} a mechanical fixture device to improve the performance of manual dissection.  This manual mosquito micro-dissection fixture (3MDF) is a simple, modular fixturing system that allows several time-consuming steps of the process to be performed concurrently on multiple mosquitoes, while also greatly simplifying the remaining per-mosquito actions performed by the human technician. We show that the time required to train an operator to perform the dissection procedure is reduced using the 3MDF, while average throughput is also increased.
    
    Though a demonstrable improvement over traditional manual methods, the 3MDF was developed only as a first step towards an end-to-end automated dissection system to enable world-wide vaccination efforts. The decision to first pursue the 3MDF proved essential to developing effective automation techniques. As a result of the 3MDF, the design of the automated system was able to become largely centered on simply automating the steps within this fixturing process: identify a mosquito, move the mosquito to a favorable position, grab the mosquito by its proboscis, place the neck in cutting blades, cut to remove the head, squeeze the exudate, and harvest the exudate. Parallel processing is a key feature of our proposed automation system for mosquito dissection. Many of these steps can be performed concurrently by a machine whereas a single technician is limited by performing them sequentially. Thus, our approach develops component systems to achieve these individual subtasks. Based on preliminary testing, we identified the rate-limiting step (as well as an absolute necessity for full automation) to be the pick-and-place of mosquito bodies. This demands accurate visual perception paired with physical precision in order to recognize a mosquito, analyze it, and best align it to remove the head. We report our work to overcome these challenges through the development of a vision-guided, robust pick-and-place robotic system for loading mosquitoes into a 3MDF-based device. As our research group works toward development of an end-to-end automated mosquito salivary gland dissection system, the demonstration of a robust pick-and-place apparatus is a key milestone in realization of that goal.

\section{Manual Mosquito Micro-dissection Fixture }
\subsection{Design}
In developing the 3MDF approach, we recognized that the fundamental problem was to align each mosquito so that the decapitation and gland extraction steps could be achieved without needless complications from extraneous mosquito parts such as legs and wings. Further, this could be accomplished with relatively simple fixturing enabling the technician to load batches of mosquitoes into cartridges aligning them so that subsequent steps could be performed in parallel. 

\begin{figure}[thb]
  \centering
  \includegraphics[width=\linewidth]{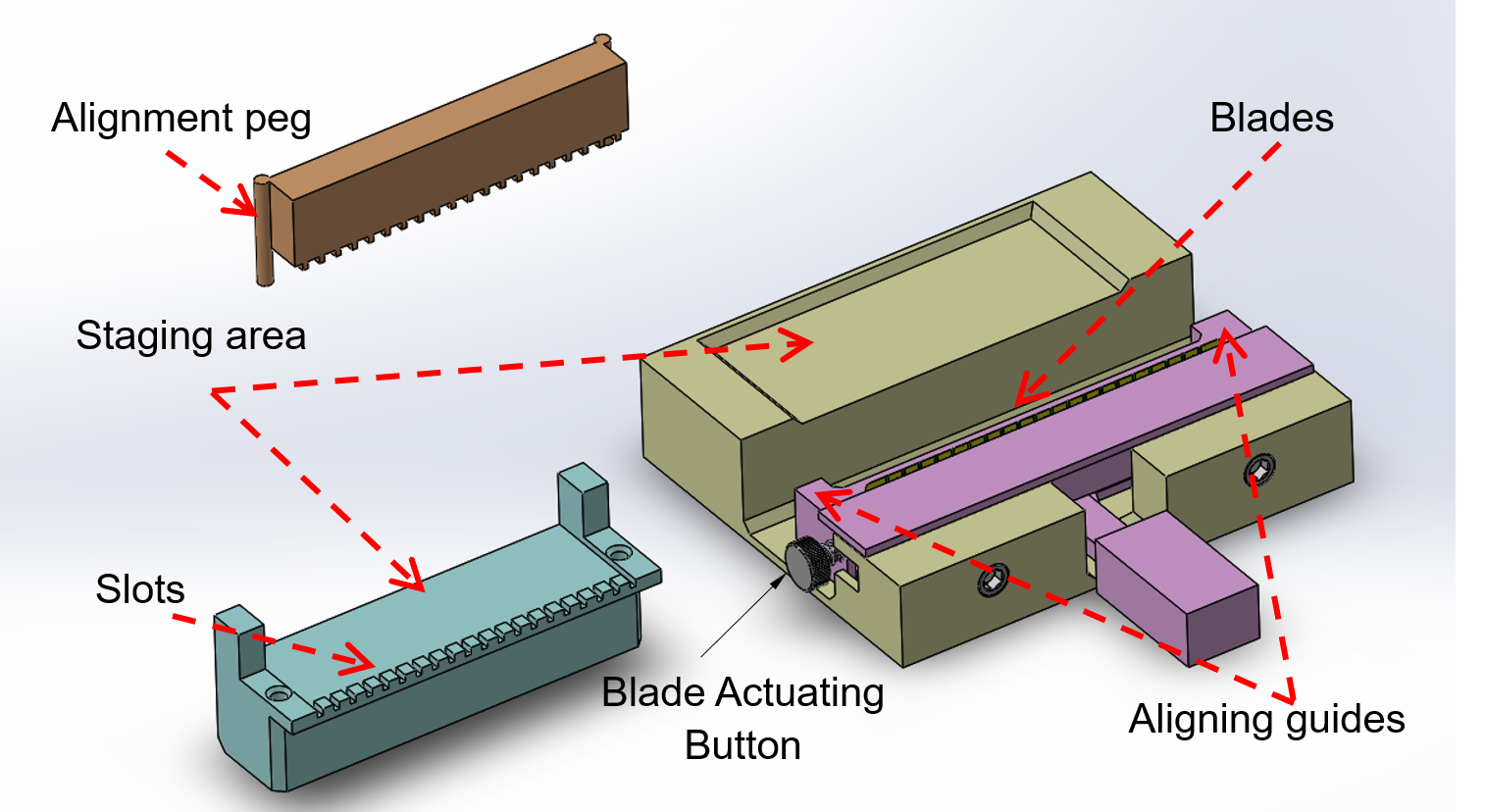}
  \caption{Design of mosquito gland extraction apparatus, including the sorting cartridge (blue), the blade assembly (pink), squeezer (brown) and staging area (tan).}
  \label{fig:cad}
\end{figure}

Our fixture design (shown in Fig. \ref{fig:cad}) consists of several modular components, including sorting cartridges, blade assembly, squeezer, and staging area. Each sorting cartridge has 20 slots allowing for the dissection of 20 mosquitoes at a time. A slot is \SI{1.25}{\milli\meter} wide, making it slightly wider than a mosquito body so that the mosquito can easily be placed in the slot but still held in position during the subsequent squeezing and harvesting stages. The slot length and depth are \SI{3}{\milli\meter} and \SI{1.5}{\milli\meter} respectively. Because the surface over which the mosquitoes are dragged must be smooth, each cartridge has a staging area made of acrylic. The staging area is \SI{71}{\milli\meter} by \SI{22}{\milli\meter}. About 30 to 40 mosquitoes can be spread out over this area which is sufficient to efficiently fill the cartridges. The sorting cartridge is removable so that when the salivary glands are extracted, they do not become trapped behind the blades. 

The blades have notches approximately \SI{0.5}{\milli\meter} wide in which the mosquito neck sits. Like hair clippers, the blades slide past one another, cutting off each mosquito head simultaneously. Because the mosquito’s neck is contacted by a sharp edge on both sides, this blade design ensures a clean cut. The blades are spring loaded so that they sit flush against the sorter cartridge. Both blades are removable, allowing for them to be easily cleaned or simply replaced. The blades are made of \SI{50}{\micro\meter} thick stainless steel and are thin enough that sharpening is not necessary. 

The squeezer is comb-like, with 20 rectangular teeth that fit into each of the slots on the sorter. To ensure the glands are extracted, the squeezer must contact each mosquito in the thorax region. Therefore, the squeezer has two round pegs that fit into matching holes in the sorter to guarantee perfect alignment. The teeth fit in the slots with very little clearance to ensure that the glands are extruded forward and do not become trapped between the side of the slot and the squeezer tooth. 

\subsection{3MDF Methods}
\begin{figure}[thb]
  \centering
  \includegraphics[width=\linewidth]{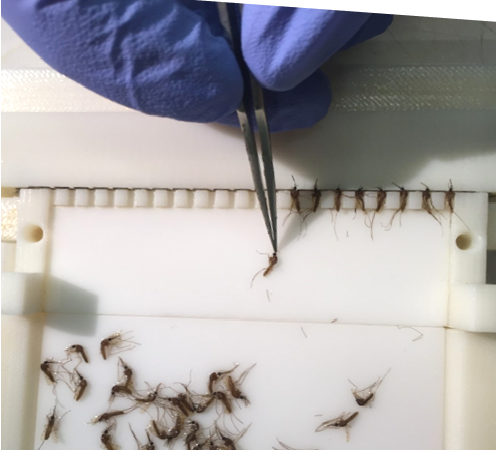}
  \caption{Sorting mosquitoes into cartridge slots. The technician grasps each mosquito by its proboscis and drags it into a slot.}
  \label{fig:hand}
\end{figure}

The workflow proceeds as follows: First, a cartridge is inserted into the apparatus so that the slots align with the blade openings, and a clump of mosquitoes is placed onto the staging area. A small amount of an aqueous medium is also placed onto the staging area. Using tweezers, the technician grasps mosquitoes one-at-a-time by the proboscis and drags the mosquito into a cartridge slot and places it so that the mosquito neck is between the clipper blades, as shown in Fig. \ref{fig:hand}. This process causes the legs and wings of the mosquito to fold back along the mosquito body, where they are constrained by the cartridge slot. This process is repeated until all cartridge slots are filled. 

Once the necks are aligned between the blades, the blades are actuated manually via a button on the side of the device (labeled in Fig. \ref{fig:cad}), enabling them to slide past one another and sever the neck of each mosquito.  Next, the sorting cartridge is removed, and another empty sorting cartridge can be inserted. 

Finally, the glands are extracted. To do this, the squeezer comb is aligned with the cartridge by placing its aligning posts into the corresponding holes in the cartridge so that the teeth on the comb rest on the mosquito thoraces. The technician then presses down to squeeze the glands out of the thoraces. The glands are ejected from the mosquito onto the flat surface in front of the blades where they can be collected by a pipettor and placed into collection tubes. 

In this study, eight untrained operators were tasked to complete the 3MDF procedure. Task completion time was tracked to determine mosquito throughput for comparison with other dissection methods.

\section{Automated System Design Concept}
\subsection{System Overview}
\begin{figure*}[thb]
      \centering
      \includegraphics[width=\linewidth]{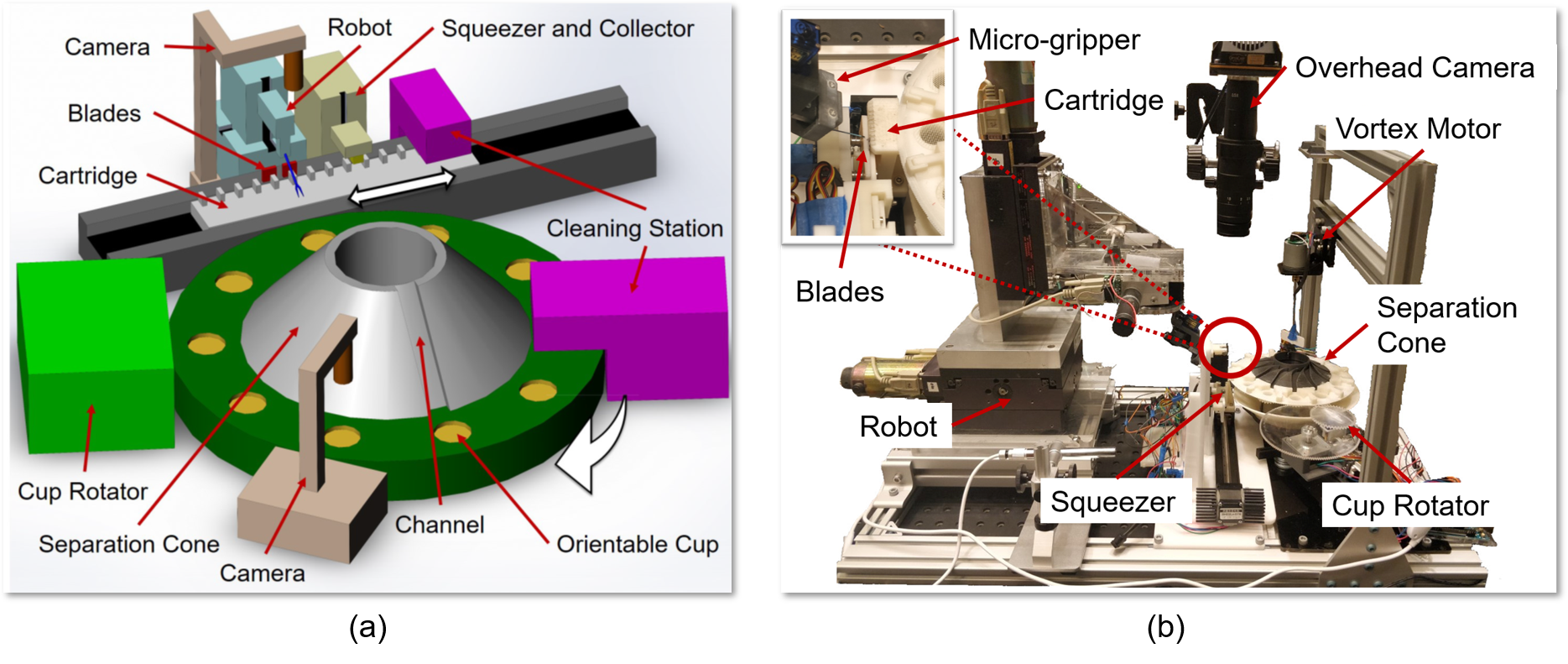}
      \caption{Automated mosquito dissection system. (a) Concept image for system. (b) Prototype realization of automated system. Background has been removed for visualization.}
      \label{fig:concept}
  \end{figure*}

The tools and methods used in the 3MDF served as a blueprint for the design of an automated mosquito dissection system, being better suited for automation compared to the traditional dissection methods. Here, we focus primarily on the robotic pick-and-place component of the fully automated mosquito micro dissection system. The larger system will ultimately take freshly sacrificed mosquitoes suspended in aqueous culture media and output a collection of mosquito exudate including PfSPZ-laden salivary glands. A concept of this dissection system and a prototype realization are provided in Fig. \ref{fig:concept}. We briefly describe this system to clarify the context of the robotic pick-and-place subsystem, which will be the middle of three primary system components. 

First, a staging apparatus will separate mosquitoes and present them one at a time to the robot. Freshly sacrificed mosquitoes sit in a basin of media beneath the system. A spinning rotor in the basin creates a vortex that will carry mosquitoes in solution to the top of a separation cone. This cone has channels in one sector down which media will flow onto a ring of orientable mesh-bottomed cups. This ring will rotationally index around the cone so that, by controlling the vortex speed and concentration of mosquitoes in the basin, the cups will on average have one mosquito on them once they pass beyond the sector of the cone with channels. At an index beyond the channel, a camera will image a single cup and a computer vision algorithm will determine if a mosquito is present. If so, at the next indexed position, the cup will be rotated to orient the mosquito so that the mosquito's proboscis will point radially outward from the ring. Finally, the ring will be rotated to an index such that the cup is aligned with a tangent linear stage that will comprise the third subsystem, a dissection assembly line. The development of the staging apparatus is described in detail in \cite{mengdi}.

The pick-and-place robot will be positioned on the other side of the linear stage and will reach over to the cup, grasp the mosquito by its proboscis and drag it onto a 3MDF-resembling cartridge attached to a linear stage. Similar to how a human technician would use the 3MDF, the robot will drag the mosquito into a slot and place the mosquito's neck into notches cut in two parallel dissection blades. An overhead camera will be used to provide computer vision feedback of this process. The blades will be actuated, cutting the head. After disposing of the mosquito's head, the robot will return to the ring which will have rotated to present a new mosquito. The linear stage will index laterally immediately after the mosquito is cut. As additional mosquito bodies are positioned on the cartridge, the linear stage will translate and expose mosquitoes to stations where a comb structure, like that in the 3MDF, will be lowered onto the mosquito to squeeze out the exudate and allow for salivary glands collection. Work on the dissection assembly line is currently ongoing within our research group. 

\subsection{Requirements}
We focus here on the robotic pick-and-place system, along with its difficult and important task of picking up mosquitoes presented on a mesh surface, and precisely placing them so that only the neck lies within the blades. In order to extract the salivary glands of the mosquito, the dissection point has to be precisely at the intersection of the head and body. If the mosquito is not placed far enough into the blade, the cut will occur on the head, leaving no passage for the exudate to be squeezed out, effectively wasting the mosquito and PfSPZ within. Because the salivary glands are located just behind the mosquito's head, placing the mosquito too far into the blade would result in some of the gland being lost in the cut, also decreasing PfSPZ yield. The mosquito neck is approximately \SI{0.3}{\milli\meter} in length, and the blades each have a thickness of \SI{50}{\micro\meter}, leaving only about \SI{200}{\micro\meter} for error. Moreover, grasping must occur only on the proboscis to prevent any damage to the body that might ruin the salivary glands or create an alternative opening from which exudate may squeeze out. The proboscis is on average \SI{2.0}{\milli\meter} long, with a diameter of approximately \SI{0.1}{\milli\meter}. 

In addition to size challenges, this procedure presents multiple difficulties not typically faced in a standard pick-and-place procedure. One of the main challenges is the mosquito-to-mosquito variation. Some of this is anatomical in origin. Each mosquito varies somewhat in size and is not axis-symmetric, meaning the alignment of the neck relative to the body depends on which side the mosquito body lies on. Further, mosquitoes are very flexible. By grabbing and pulling the mosquito from the proboscis, the body tends to straighten out in time for placement, but first, the mosquito must be identified and grasped from a variety of twisted, compressed, or otherwise contorted orientations. While upstream processes are expected to align the mosquito's proboscis within 15 degrees of an ideal orientation for the robot to grab, the mosquito can still be located anywhere on the cup and must be grasped accordingly. Because of its length, the proboscis can still be grabbed even if there is some error in the robot positioning, or computer vision targeting. However, this, combined with the general variability in proboscis lengths, means that the offset between the robot's grasp point and the mosquito's neck is not consistent trial-to-trial. As such, it is not enough to program a sequence of robot movements; these challenges necessitate adaptive automation. Based on visualization of the mosquito’s anatomy and its grasping location, the robot must perform customized movements to successfully grasp and place the mosquito.

\subsection{Experimental Setup}
\begin{figure}[thb]
      \centering
      \includegraphics[width=\linewidth]{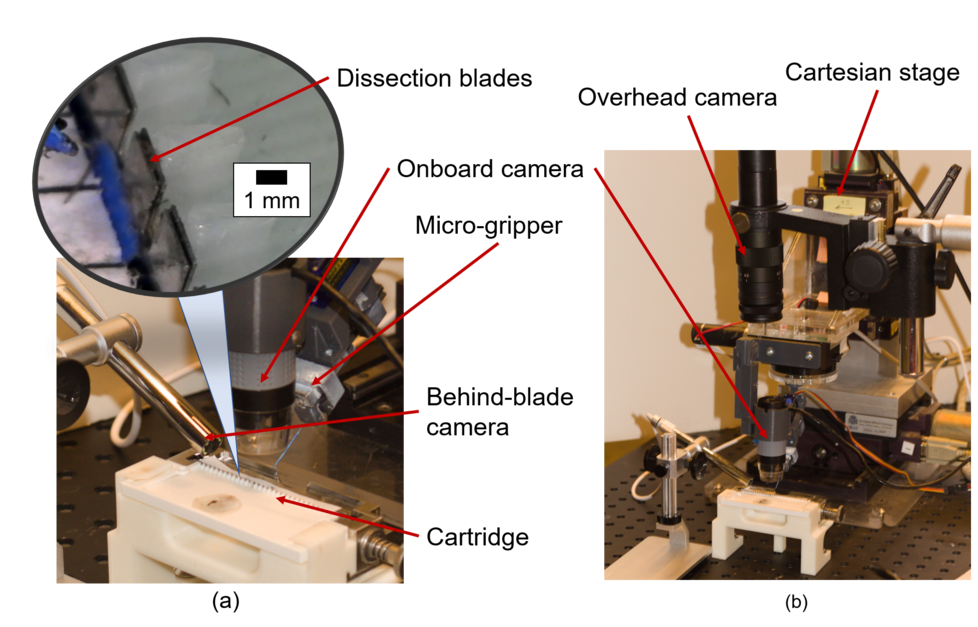}
      \caption{Experimental Setup. A close-up image of dissection blades and cartridge is inset with length scale for reference.}
      \label{fig:setup}
  \end{figure}

\begin{figure}[thb]
      \centering
      \includegraphics[width=\linewidth]{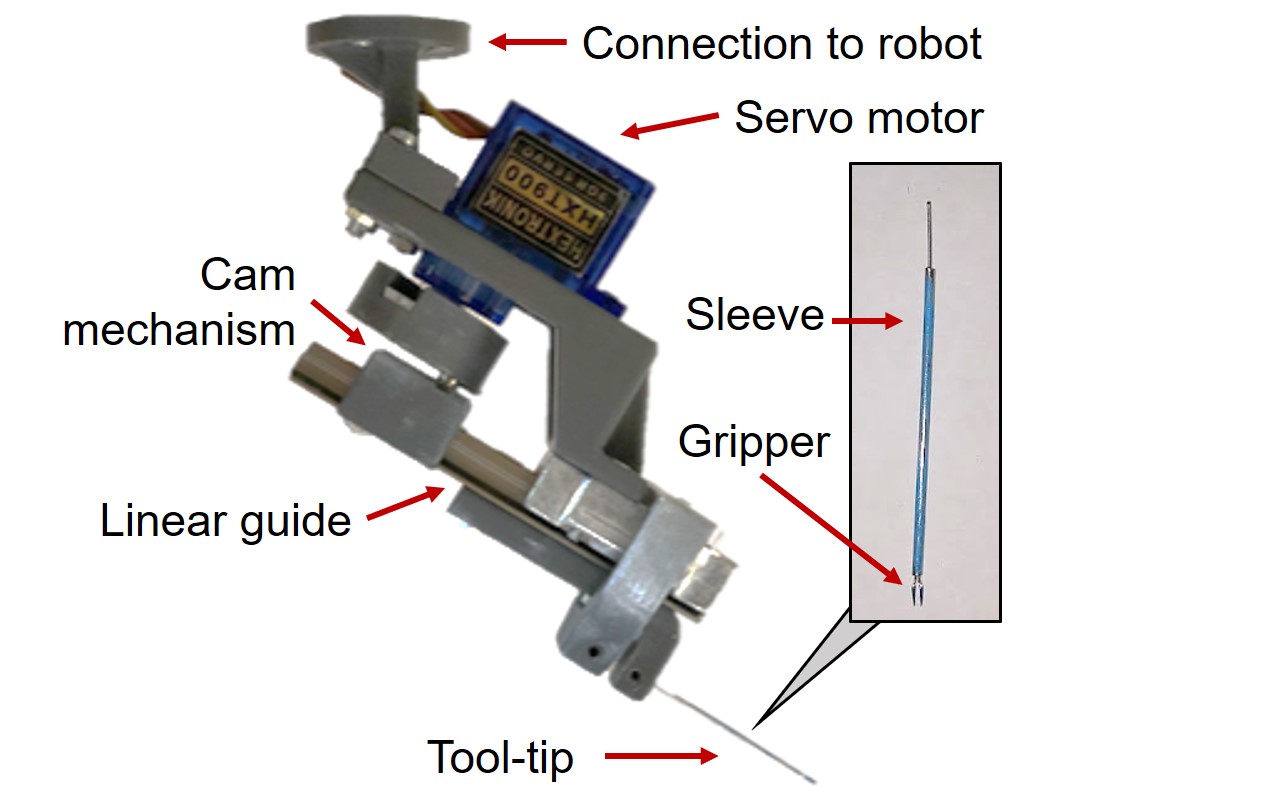}
      \caption{Custom-designed micro-gripper used to grasp mosquitoes.}
      \label{fig:gripper}
\end{figure}

The robot used in this procedure is a 4-DOF, linear stage robot by New England Affiliated Technologies, Lawrence, MA (Fig. \ref{fig:setup}). A dual-axis X-Y table is used as the base for the robot, onto which a Z axis is mounted orthogonally (NEAT: XYR-6060 and NEAT: LM-400 respectively). The robot also has a rotary axis which is not used in this study. Each axis is driven by a 12V DC servo motor, with a leadscrew, has a travel of \SI{100}{\milli\meter}, and is coupled with an incremental encoder. The positioning resolution of these axes was measured with a dial indicator to be approximately \SI{10}{\micro \meter}. The entire assembly is mounted to an optical table. Robot motion is driven by a Galil controller (DMC-4143), interfaced to a Linux computer by ethernet connection. Attached to the robot is a custom-designed micro-gripper mechanism visualized in Fig. \ref{fig:gripper}. A cam mechanism attached to a HexTronik HXT900 servo motor drives the rail of a linear guide within its carriage, causing the tooltip to open and close. The tooltip of the micro-gripper is adapted from an Alcon Grieshaber retinal surgical forceps. Movement of the linear guide rail extends or retracts a sleeve over normally-open gripper jaws. The micro-gripper is controlled by sending position commands to the servo motor via USB serial communication from the computer to an Arduino Uno microprocessor.
 
Mosquitoes are staged for dissection on a modified 3MDF device that is also mounted to an optical table. The 3MDF cartridge is modified to have a hole \SI{23}{\milli\meter} away from the blades in which is placed a \SI{20}{\milli\meter} diameter cup that matches the one used in the upstream staging apparatus. This cup is covered with a nylon \SI{750}{\micro \meter} mesh that is used for media drainage in that apparatus. The mosquito is dragged into a slot in the 3MDF cartridge and placed into the 3MDF blades. These are two \SI{50}{\micro \meter} thick stainless steel blades with \SI{0.5}{\milli\meter} wide by \SI{1.0}{\milli\meter} deep notches cut in them to match the midpoint of the slots. The closest blade to the cartridge is stationary while the further blade can be manually actuated by pressing a button on the side of the device. This action causes the mosquito neck to be caught between the two blades and cut.

The setup also includes three cameras (Fig. \ref{fig:setup}). An overhead microscopic camera (OptixCam Summit D3K2-5) with an Omano OM-10K zoom lens is used to capture a complete view of the workspace and is used by the computer vision to identify a mosquito's presence and approximate location. A second camera (Plugable USB Microscope Camera), is mounted on the robot and is used to identify the location of the mosquito's body parts for accurate picking and placing. We refer to these as the overhead and onboard cameras respectively. A third camera (Opti-Tekscope USB Microscope Camera) is placed to the side and rear of the setup so that its visual field is in line with the blades. This camera is not necessary for system operation and is only used to visualize placement to evaluate trial success.

\begin{figure}[thb]
  \centering
  \includegraphics[width=\linewidth]{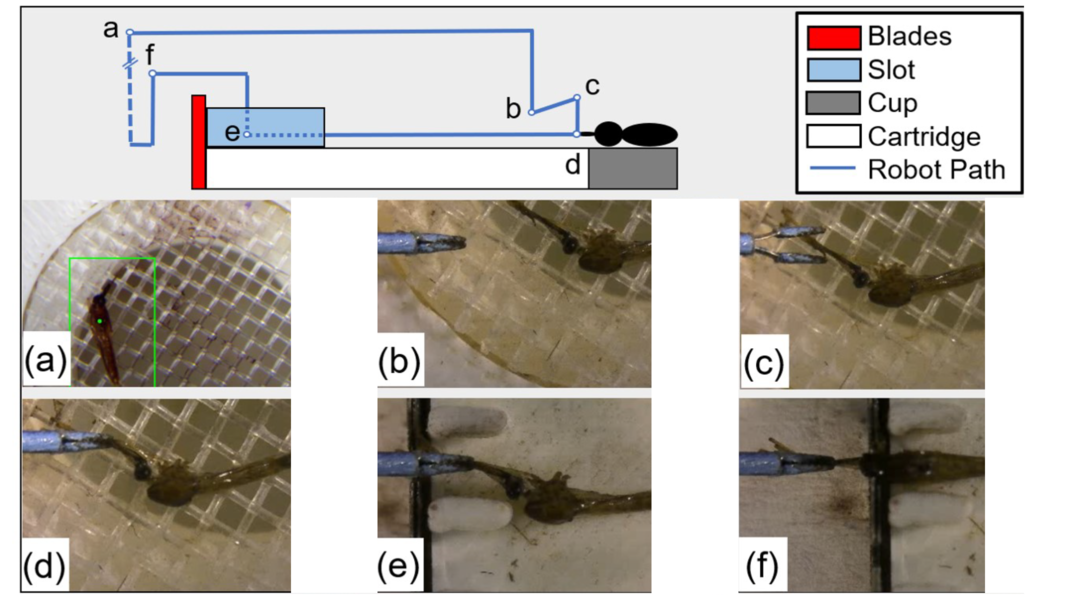}
  \caption{Side view of robot path and related representative images captured by the vision system. (a) Image captured from overhead camera showing bounding box of detected mosquito. (b) Image captured from onboard camera to determine proboscis centroid. (c) Image captured from onboard camera before grasping. (d) Image captured from onboard camera immediately after grasping. (e) Image of the mosquito taken used to calculate head-to-tooltip offset. (f) Image after aligning the mosquito neck with the blades.}
  \label{fig:combined}
\end{figure}

The automated procedure uses the overhead and onboard cameras to guide the robot's motion. The procedure consists of three stages. In the first stage, an image of the entire workspace is captured using the overhead camera. This image is converted to HSV space and the mosquito is segmented out. Next, a bounding box is fit to this region and a weighted centroid is calculated for the mosquito, as shown in Fig. \ref{fig:combined}(a). The tooltip is moved near the mosquito, to a position where the onboard camera, due to its close proximity, is able to capture an image of the mosquito with more features and details.

In the second stage, a computer vision algorithm identifies the mosquito's proboscis in the detailed onboard camera image shown in Fig. \ref{fig:combined}(b). The tooltip is moved to a point above the centroid of the proboscis (Fig. \ref{fig:combined}(c)), which is used as the grasp location for the mosquito (Fig. \ref{fig:combined}(d)). Finally, the robot drags the mosquito to an empty slot on the cartridge near the blades.

In the third stage, the onboard camera captures a final image shown in Fig. \ref{fig:combined}(e) with the tooltip in view to detect the mosquito head-to-tooltip offset. The robot uses this offset value to position the mosquito with its neck between the dissection blades (Fig. \ref{fig:combined}(f)). Our group is also investigating the use of a keypoint-based computer vision approach \cite{hongtao}, which was not used in this study.

 \subsection{Calibration} 
To relate the robot and camera coordinate systems, we use a two stage calibration process. In the first stage, the tooltip of the micro-gripper is located in the overhead camera frame. The tool is segmented from the background in HSV space using Otsu's binarization \cite{otsu}, and contour identification is used to detect the tool. The lowest point of the tool contour is then used as the tooltip. In the second stage of calibration, the robot is moved through the camera space across a grid of points. The tooltip is detected and recorded at each position. The resulting grid of points from both coordinate systems are used as the inputs for a Bernstein polynomial fitting routine as performed in \cite{bernstein}. In this routine, two fourth degree polynomials are fit to create a mapping from the camera coordinates to the robot's encoder coordinates. Such a polynomial fitting method also compensates for radial and aspherical lens distortions. 

The tooltip does not move with respect to the overhead camera, so the polynomial fitting method described above could not be used to calibrate this camera. Instead, a pre-calibrated grid of a resolution \SI{5}{\milli\meter} x \SI{5}{\milli\meter} was placed in the background. The robot was then moved by a known distance along each axis, and images were captured before and after motion. The pixel motion of the grid was calibrated to the corresponding change in robot encoder counts.

The location of the cartridge and blades in robot coordinates were determined using a shim. The robot was slowly advanced until the robot held the shim firmly to the surface of interest and encoder counts at this location were used as a reference.

\section{Mosquito Localization and Segmentation}
The vision system consists of two cameras: the overhead camera and the onboard camera, as shown in Fig. \ref{fig:setup}. The two cameras assist in the task of localizing the mosquito position and orientation within the robot workspace and then identifying the proboscis grasp location. This is done in multiple stages to pick and place the mosquito accurately, as shown in Fig. \ref{fig:combined}.

\subsection{Approximate localization using overhead camera}
The first stage consists of locating the mosquito in the overhead camera frame, captured at a high resolution (2560x1922 px). This is done with a series of operations on the obtained image: 
\begin{enumerate}
    \item Gamma Correction is performed to remove high intensity details of image, and to increase the contrast of the lower intensity mosquito region in the image.
    \item The image is then cropped to remove blades and the fixture at static locations in the image.
    \item The image is then converted to HSV space, and the saturation channel is selected as the most representative image for segmentation.
    \item Gaussian Blur ($\sigma = 15px, K=(15,15)$) is then applied to the saturation channel to reduce noise and false edges within the image. 
    \item Next, Otsu's \cite{otsu} binarization is used to obtain optimal upper and lower bounds for thresholding, maximizing the variance between groups. This leads to an approximate binary mask representing the mosquito within the image.
    \item Morphological opening operation is then performed on the binary mask to remove smaller objects from the foreground.
    \item Gaussian blurring ($\sigma = 15px, K=(15,15)$) is then applied to further smoothen the image, and remove any noise in the mask.
    \item Morphological erosion ($K=(30,30)$) is then performed to particularly smoothen the contours of the mask.
    \item Connected component analysis is then used to select the largest regions within the mask image, above an area threshold (5\% of image area).
    \item The largest regions of the mask obtained and its bounding boxes are then used as approximate locations of mosquitoes from the overhead camera. The bounding box and centroid of the mosquito is as shown in Fig. \ref{fig:combined} (a) in green.
\end{enumerate}

The bounding box obtained is used for pick procedure to move the gripper to initial location.

\subsection{Accurate location using onboard camera}
\label{subsect:onboard_cam}
In the second stage, the onboard camera image is used. The mosquito needs to be grasped on its proboscis for a successful grasp so no damage is made to the body or the salivary glands within the thorax. To dissect the mosquito, the neck needs to be accurately placed within the 3MDF device. For this purpose, a deep convolutional neural network is used to identify different anatomical regions of the mosquito such as proboscis, head, or body from a given image as a dense pixelwise mask. This mask is then used to infer the grasp point and the dissection point accurately in stage 2 and 3.

\begin{figure*}[thb]
      \centering
      \includegraphics[width=\textwidth]{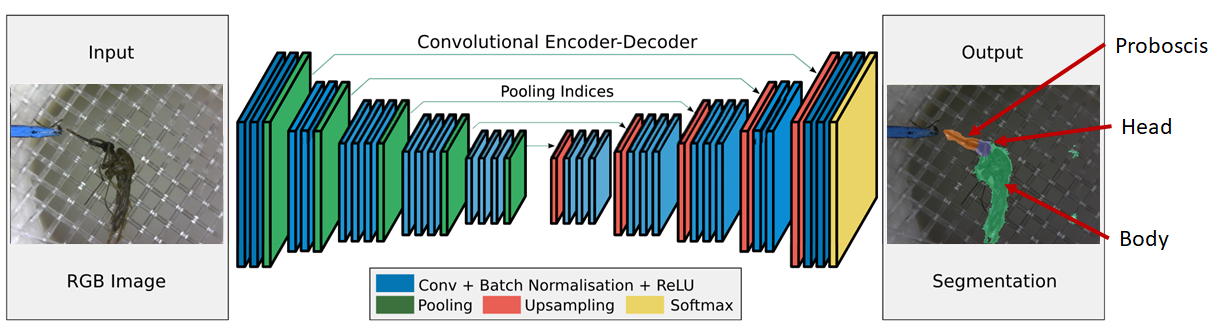}
    \caption{Network architecture for segmentation along with overlaid segmentation mask on image showing the detected proboscis, head, body and the background.}
    \label{fig:vgg}
\end{figure*}

To train the network on such a novel task, a dataset of 185 images (1600x1200 px) of mosquitoes in various positions and orientations were captured from the onboard camera. This dataset was then annotated manually with dense pixel labels for the three regions: proboscis, head, and body. The images and labels were then downsampled to a resolution of 300x400 px for faster training. To leverage the power of previous work in image classification in computer vision, a SegNet model \cite{segnet} pretrained with VGG-16\cite{vgg16} weights was used. This enables the initial layers of the network to function as generalizable feature detectors, whereas the final layers are fine-tuned by transfer learning specifically for this task.

To generalize the network weights and generate a larger training dataset, rigorous data augmentation was used during the network training phase. Since the mosquitoes can present in any orientation or position, random rotation between $(-\pi,\pi)$ was applied along with random X-Y reflections. The image-label pairs were then randomly translated along both X and Y axes by \rpm100px. Finally the pairs were also scaled randomly between scaling factors $(0.75,1.25)$ to compensate for variability in sizes of mosquitoes. Both Generalized dice loss \cite{generalDice} and multi-class cross entropy loss were tested as the loss functions, however weighted multi-class cross entropy loss yielded better results. Since the proboscis occupies a much smaller region within the image in comparison to the background, each class was weighted by the inverse of area of each region in training labels. The loss function was modified accordingly.

The model weights were then fine-tuned on this dataset for 1500 epochs, after which the model converged and no significant changes in accuracy were observed. During inference, the onboard camera only encounters cases where one mosquito is in the image at any instance. To enforce this prior knowledge, using connected component analysis, the largest regions within each label class were then selected as the mask. All the other regions were assigned as background class. Another prior condition enforced during inference was the edge connectivity between proboscis to head, and head to body. This was done by dilating both relevant classes and using intersection of images as the edge. The centroid of this edge was then used as the dissection point between head and body.

\section{Experimental Methods}
\subsection{Study Design}
Testing was performed to investigate the efficacy of the designed system to pick up a mosquito and place that mosquito within blades that can remove the insect's head. The process was performed on 50 non-infected \textit{Anopheles} mosquitoes. Prior to testing, the mosquitoes were kept in an airtight, refrigerated container of phosphate-buffered saline solution (PBS) following sacrifice one day prior. Functioning as a unit test for this subsystem within the eventual automated mosquito dissection system, only the grasping and subsequent positioning of the mosquitoes by the robot were considered for trial success or failure. All actions of the system during the test were performed autonomously with feedback from computer vision, and a manual cut was performed at the end of each trial to facilitate determination of trial success.

\subsection{Pick Procedure}
The experimental procedure is demonstrated in Fig. \ref{fig:combined}. A mosquito is removed from the PBS solution by its proboscis with tweezers and placed anywhere on a circular mesh cup of radius \SI{10}{\milli\meter} with its center placed \SI{23}{\milli\meter} away from the blades as measured on the central axis of the cartridge slot. The mosquito was placed so that the proboscis was positioned forward toward the blades, pointing within 45 degrees this line. One such placement is shown in Fig. \ref{fig:combined}(a). These conditions were chosen to mimic the worst case expected from the upstream mosquito-staging apparatus that this process will later be integrated with. To further match the expected results of this upstream process, no further attempt at standardization of mosquito starting position were made (e.g. what side the mosquito was lying on, relative straightness of legs). The micro-gripper tooltip begins the trial at a location away from the cup and \SI{3.5}{\milli\meter} above the cartridge surface. 

A bounding box around the mosquito is identified by computer vision in an image from the overhead camera, and the robot moves to a point \SI{5.0}{\milli\meter} in front of the centroid of that region (Fig. \ref{fig:combined}(b)). This brings the mosquito into view of the onboard camera without placing the gripper over top of the mosquito body. By lowering \SI{3.0}{\milli\meter} toward the mesh surface, the mosquito is brought into focus. The centroid of the proboscis region is identified and the robot moves the gripper to a location \SI{2.0}{\milli\meter} above this this point (Fig \ref{fig:combined}(c)), and then drops down to the mesh surface and the gripper is closed to grab the proboscis (Fig \ref{fig:combined}(d)). 

The robot lifts up \SI{0.8}{\milli\meter} and drags the mosquito to a position \SI{1.5}{\milli\meter} from the blades (Fig. \ref{fig:combined}(e)). Here, an image from the onboard camera is again analyzed by the computer vision system. This task serves two functions, to confirm successful grasping of the mosquito, and to determine more accurately where on the proboscis the gripper has grabbed. The trial is considered a successful demonstration of grasping if the mosquito is visualized as being held within the micro-gripper at this point.

 \subsection{Place Procedure}
 The vision system provides the location of the proximal end of the proboscis, where it attaches to the mosquito's head. This location is transformed into robot coordinates and a head-to-tooltip offset is determined by subtracting it from the current encoder values. Only the offset in line with the cartridge grooves (a horizontal offset in Fig. \ref{fig:combined}(e)) is considered. The robot then executes another set of programmed movements. The robot raises the mosquito head \SI{1.3}{\milli\meter} and moves forward a nominal distance to clear the blades plus the offset, such that the mosquito's neck should be right above the blades (Fig. \ref{fig:combined}(f)). Then the tooltip moves down \SI{3.0}{\milli\meter}, placing the neck within the notch of the blades if properly aligned. At this point, another subsystem of the automated mosquito dissection system would actuate the blades to cut the head and further process the mosquito. In this unit test, the blade is manually actuated. The test is considered a successful placement if the mosquito's neck is placed into the notch of the dissection blades such that the head could be removed. As a final step of the process, the robot pulls away from the blade, moving the head, if still in its grasp, to a location where it can be cleaned off with a modest jet of air or fluid that does not disturb the tooltip calibration. Video footage from all three cameras is recorded throughout and saved for analysis. The commanded speed of each robot axis was \SI[per-mode=symbol]{12.5}{\milli\meter\per\second}, chosen somewhat conservatively compared to the robot's top speed to minimize overshoot. The speed was decreased to \SI[per-mode=symbol]{2.5}{\milli\meter\per\second} when lowering the mosquito neck into the blades, reducing the inertia of the mosquito and thus the tendency to pivot or flip over the blades rather than settle between them.
 
\section{Results}
\subsection{Training and Performance Comparisons}
\begin{figure}[thb]
      \centering
      \includegraphics[width=\linewidth]{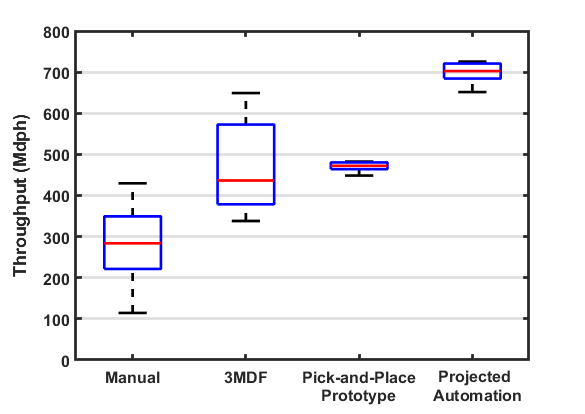}
      \caption{Mosquito throughput based on dissection method. Shown are throughput rates in mosquitoes dissected per hour (Mdph). The boxes represent $25^{th}$ to $75^{th}$ percentiles of data, and the red lines represent median values. The traditional manual process is from internal production data from Sanaria with trained operators. The 3MDF data represents the results in Table \ref{table:samms_data}. The pick-and-place throughput data is estimated based on cycle times of the robotic system in this test. The projected automated system throughput is added to show the estimated throughput with optimized robotic movement, calculated by removing 2.5 seconds from each cycle time in this test (see section \ref{sect:discussion}).}
      \label{fig:comb_results}
  \end{figure}

Using the 3MDF, untrained operators demonstrated high production throughput rates. Even in their very first trials, operator times ranged from 338-649 mosquitoes per hour (Table \ref{table:samms_data}). The average throughput was 470 mosquitoes dissected per hour (Mdph). After this test, trainees continued their course over the next several days. Entrainment, as defined by Sanaria's procedures, occurred over as little as 3 trials. The number of sessions required to complete training was fairly consistent, taking one day (3–10 trials over one hour) to 3 days, compared to the manual process which averaged 16 weeks with 1-3 one-hour sessions per week. This represents at least a five-fold reduction in average training time. Every trainee was able to achieve a total mosquito alignment and decapitation time for a 20 mosquito cartridge of less than 1.5 minutes and gland collection time of 0.5-1.0 minute, similar to operator 6, (Table \ref{table:samms_data}) and a total output of 600 mosquitoes per hour. In comparison, the dissection rate by the traditional manual methods performed by trained operators averaged around 290 Mdph with a wide range in individual operator capabilities (260 to 430 Mdph) as shown in Fig. \ref{fig:comb_results}. 

The automated pick-and-place task, which our preliminary tests indicated would be the rate-limiting step of the end-to-end automated dissection system, took on average 7.5 seconds to complete one cycle in this test (Fig. \ref{fig:comb_results}). This consisted of 7.34 seconds for robot movement and 0.16 seconds for image processing on average. This corresponds to a rate of approximately 470 Mdph, with a small variability (standard deviation of 11 Mdph) due to variable placement resulting in different distances over which the mosquito was dragged in a given cycle. There remains room for substantial optimization of the robot trajectory in future design iterations. The right-most box in Fig. \ref{fig:comb_results} represents the theoretical throughput associated with removing 2.5 seconds from the robot movement for each cycle, which we predict would be possible without mechanical changes (see Section \ref{sect:discussion}).

\begin{table}
\caption{Production Rates for Untrained Operators using 3MDF}
\label{table:samms_data}
\begin{center}
\begin{tabular}{|c c c c c c c c c c|}
 \hline
  & 1 & 2 & 3 & 4 & 5 & 6 & 7 & 8 & Avg. \\
 \hline
  (A) & 2.4 & 2.5 & 2.3 & 2.5 & 1.5 & 1.3 & 1.1 & 1.2 & 1.9\\ 
 \hline
 (B) & 0.7 & 0.8 & 0.5 & 1.1 & 1.2 & 0.7 & 1.1 & 0.7 & 0.8\\
 \hline
 Total & 3.1 & 3.3 & 2.8 & 3.6 & 2.7 & 2.0 & 2.2 & 1.9 & 2.7 \\
 \hline
 Rate & 393 & 364 & 429 & 338 & 444 & 600 & 545 & 649 & 470\\
 \hline
\end{tabular}
\end{center}
(A) Time taken (in min.) to align 20 mosquitoes; (B) Time taken (in min.) for gland extrusion and collection. Total time for 20 mosquitoes (in min.), and rate is mosquito throughput per hour. Eight untrained operators were tested as a part of this study.
\end{table}

\subsection{Mosquito Localization and Segmentation}
\begin{figure}
  \centering
  \includegraphics[width=\linewidth]{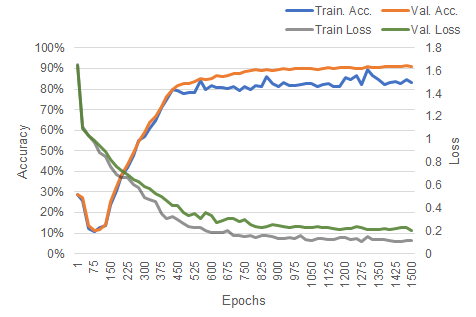}
\caption{Network training accuracy and losses. The network converged after approx. 1500 epochs taking 195 mins, with a training accuracy of 84.0\% and validation accuracy of 91.8\%.}
\label{fig:losses}
\end{figure}

\begin{figure}
  \centering
  \includegraphics[width=\linewidth]{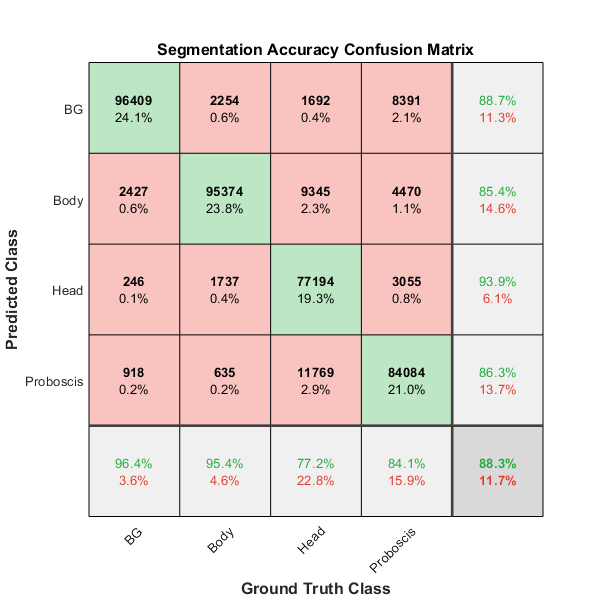}
\caption{Normalized pixelwise classification accuracy, highlighted according to classification accuracy. The overall network accuracy across all classes is 88.3\%.}
\label{fig:confusion}
\end{figure}

The segmentation network was trained for 1500 epochs and both training and validation losses converged as shown in Fig. \ref{fig:losses}. The network accuracy was measured using multiple metrics as defined and described in \cite{segmmetrics}, including pixelwise accuracy and weighted intersection-of-union(w-IoU). Since the classes are highly imbalanced with proboscis and the head being most important, w-IoU is selected as a metric to combine accuracy and class weight. Note that the validation accuracy is always higher than the training accuracy since we use rigorous data augmentation during training phase of the network. The global pixelwise accuracy across all classes was 88.3\%, however this includes the background as a class. Only the proboscis and head are of particular interest and the accuracy obtained for both were 0.7719 and 0.8408 respectively. The confusion matrix for class accuracy is as shown in Fig. \ref{fig:confusion}. The prediction improved further after performing the edge-connectivity filtering and post-processing steps as described in Section \ref{subsect:onboard_cam}. An example of segmentation predicted by the network is as shown in Fig. \ref{fig:vgg}, as the output of network.
    
\subsection{Automated Pick-and-Place Results}
\begin{figure}[thb]
      \centering
      \includegraphics[width=\linewidth]{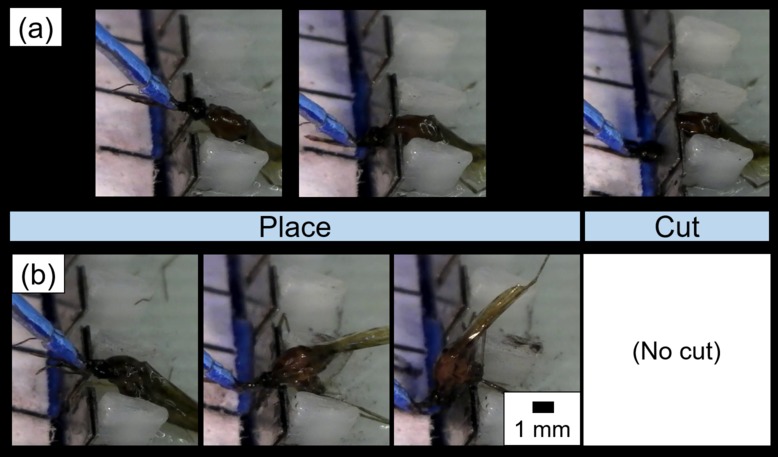}
      \caption{A demonstration of mosquito placement. (a) A mosquito being accurately placed with neck between the blades. (b) Inaccurate placement of the mosquito, resulting in the body pivoting over the blades when the proboscis is pulled down by the gripper, rather than the neck settling between the blades.}
      \label{fig:results}
  \end{figure}

Throughout the automated experiment, there were no issues moving to a mosquito's location, grasping it by the proboscis, or dragging it on the surface of the cartridge. All 50 (100\%) of the mosquitoes were observed with the proboscis grasped by the micro-gripper during the second vision check (Fig. \ref{fig:combined}(e)). Of these 50 mosquitoes, 45 (90\%) were placed such that their necks were aligned correctly within the blades. Placement was considered successful if the alignment allowed the blades to cut the neck such that the head could be fully removed.  The results were confirmed post-test with close-up video taken of the blades during the placement and cutting steps. An example of a mosquito being accurately placed is provided in Fig. \ref{fig:results}(a).

The five mosquitoes that were not accurately placed exhibited similar behavior, flipping over the blades when pulled down by the robot. This action is demonstrated in Fig. \ref{fig:results}(b). In these cases, the mosquitoes appear to collide with either the slot walls or the blades. That collision point acts as fulcrum, causing the downward motion of the robot to flip them over the blades, rather than pull the neck into the notches. We were unable to correlate this behavior with any other variable including initial mosquito orientation, grasp location of the proboscis, trial number, or a qualitative assessment of the computer vision's head-to-gripper offset estimation.

\section{Discussion}
\label{sect:discussion}
The immediate goal was to develop production fixtures and methods that could significantly improve the existing manual process while providing the experience necessary to develop a fully automated mosquito dissection and gland extraction process. Accordingly, we have developed prototypes for robust mosquito alignment, decapitation and gland extraction, and demonstrated highly successful performance of a robotic pick-and-place subsystem. 

Gland extraction from mosquito thoraces is vastly simplified and deskilled using the 3MDF as indicated by improved training capability. The training time to qualify individual operators for mosquito processing is radically reduced compared to manual dissection while simultaneously doubling throughput. While not as easily scaled or as consistent as an automated solution, the 3MDF appears to provide a substantial interim improvement, while simultaneously serving as the task template for an end-to-end automated system. This is evidenced by the successful development of an automated system to perform the pick-and-place task from the 3MDF method. The demonstration of a simplified dissection procedure, coupled with automation of a key component step, provides confidence in our approach as we work to automate all 3MDF procedure tasks and move towards fully automated mosquito dissection for malaria vaccine production.

Developing the 3MDF required considerable experimentation with alternative designs before we converged the embodiment described here, and we will discuss these experiences briefly. The approach of grasping each mosquito’s proboscis and dragging it across a surface lubricated with aqueous culture medium into an appropriate cartridge slot was very successful from the beginning. We explored several different decapitation methods and apparatus before settling on the clipper blades arrangement described above. This method reliably severs the necks with minimum displacement of the mosquito bodies, unlike single-blade methods. Also, technicians do not find it difficult to guide mosquitoes so that the necks are properly positioned between the blades. Similarly, we explored several alternative approaches for salivary gland extraction and collection before finally settling on our current comb squeezer and pipette collection approach. In particular, we found that having a fairly tight fit between the squeezer comb teeth and the cartridge slots ensures that the gland material is extruded to the front surface of the cartridge. At this point, the glands tend to stick to the front surface, and the suction device can easily gather them, as well as any that have slid down to the bottom of the cartridge. 

The robotic pick-and-place subsystem demonstrated highly successful results in this unit test. With no failures in grasping or moving the mosquito, the micro-gripper and computer vision methods were shown to be adequately designed for the task.  The ability of the system to achieve these promising results indicates that the computer vision system was effective at providing appropriate adaptations to robot movement. Although we are still working to improve the system, the 90\% success rate from this study is very encouraging, especially considering the challenges presented by this non-standard pick-and-place task. It is not surprising that the placing task would prove more difficult than grasping the mosquitoes as it requires more accuracy. In order to move the mosquito, the robot can grasp anywhere on the length of the proboscis. Placing the mosquito's neck between the blades requires more precision and any inconsistencies in grasping, deformation, and anatomy must be accommodated in this step.
In the few cases where adequate placement was not achieved, the mosquitoes were observed to flip over the blades about a contact point with either the blades or the cartridge. This behavior occurred both when the neck appeared to be misaligned with the blade notches as well as in cases when the alignment appeared adequate. When there was neck misalignment, either the head or body of the mosquito, which are wider than the notch within the blades, contacted the top of the blades and caused the mosquito to flip over when the robot pulled the proboscis downward. Our work to better determine the tooltip-to-head offset should improve the accuracy of alignment. As the robot holds the proboscis above the cartridge surface, its length is foreshortened in the top-down view provided by the onboard camera. A better estimation of the offset may be obtained geometrically or from a side-view camera where the proboscis profile should not be distorted. We will also target further improvement through mechanical changes to the blade and cartridge geometries to better guide the mosquito neck into position even in cases of small errors in robot positioning. These modifications should also address the situations in which alignment appeared adequate by video observation.

The robot movements combined took 7.33 seconds on average in this test. The computer vision steps have been optimized to run on a GPU at a rate of 0.16 seconds on average per image. As the first vision step will happen in parallel with later robot movements this time can be excluded from the total cycle time count. Without hardware changes, we estimate up to 2.5 additional seconds could reasonably be reduced, leading to an average rate around 700 mosquitoes per hour. These time savings would be achieved by means of movement optimizations such a less conservative speed, both throughout and specifically while inserting the neck between the blades, and movement of the home point of the robot closer to the cup. The final system is also expected to use a smaller, lighter robot, which should achieve faster accelerations and may allow for further time improvement through increased nominal speed. Still, automation of the dissection process can be beneficial even before large increases in throughput can be realized as the consistencies associated with lower variability (e.g. in SPZ yield, cost) should ease downstream processes in vaccine production. 

\section{Conclusion}
We have presented the design and workflow for a manual mosquito micro-dissection system that addresses a significant problem in the production of a malaria vaccine, as well as demonstrates the automation of the challenging and non-standard pick-and-place task involved in mosquito dissection. Our 3MDF design is currently being refined to improve multi-user and long-term use and to ensure materials compliance for manufacturing under current good manufacturing practices (cGMPs) as specified by the FDA. We are in the process of implementing a cGMP-compliant version of the 3MDF to extract mosquito salivary glands in the production scheme for phase III clinical trials. 

Aside from its near-term value in increasing productivity while reducing operator training time, the 3MDF also serves as a blueprint in the development of a fully automated dissection system. Since the system design is modular, and each component is shown to work manually, we can integrate additional automated components as they are designed. In the near term, we will focus on automating the downstream 3MDF processes (head removal, squeezing, and collection) and integrating them with our vision-guided robotic pick-and-place system, while also further refining our vision algorithms. We will then move on to development of an improved automatic feeding process. As we move forward, additional challenges will include optimizing throughput and implementing further error detection and recovery actions. 
The successful demonstration of the robotic pick-and-place component of this system represents a major milestone in our effort to automate the malaria vaccine production process. Although the system is highly specialized, it could be adapted to develop treatment for other mosquito-based diseases. Our systems approach and many of the basic techniques might also be applied to other image-guided automation applications for handling of small, delicate, or variable materials. We hope that our experience reinforces to the reader the key role that manual process refinement, such as the development of the 3MDF, plays in the development of automated technology. 

%



\section*{Acknowledgments}
This work was supported in part by NIH SBIR grants R43AI112165 and R44AI134500. Additionally, H. Phalen is supported by the National Science Foundation Graduate Research Fellowship under Grant No. DGE-1746891.

\ifCLASSOPTIONcaptionsoff
  \newpage
\fi


\def\UrlBreaks{\do\/\do-}  

\bibliographystyle{IEEEtran}
\bibliography{IEEEabrv,references}

%



%

\begin{IEEEbiography}[{\includegraphics[width=1in,height=1.25in,clip,keepaspectratio]{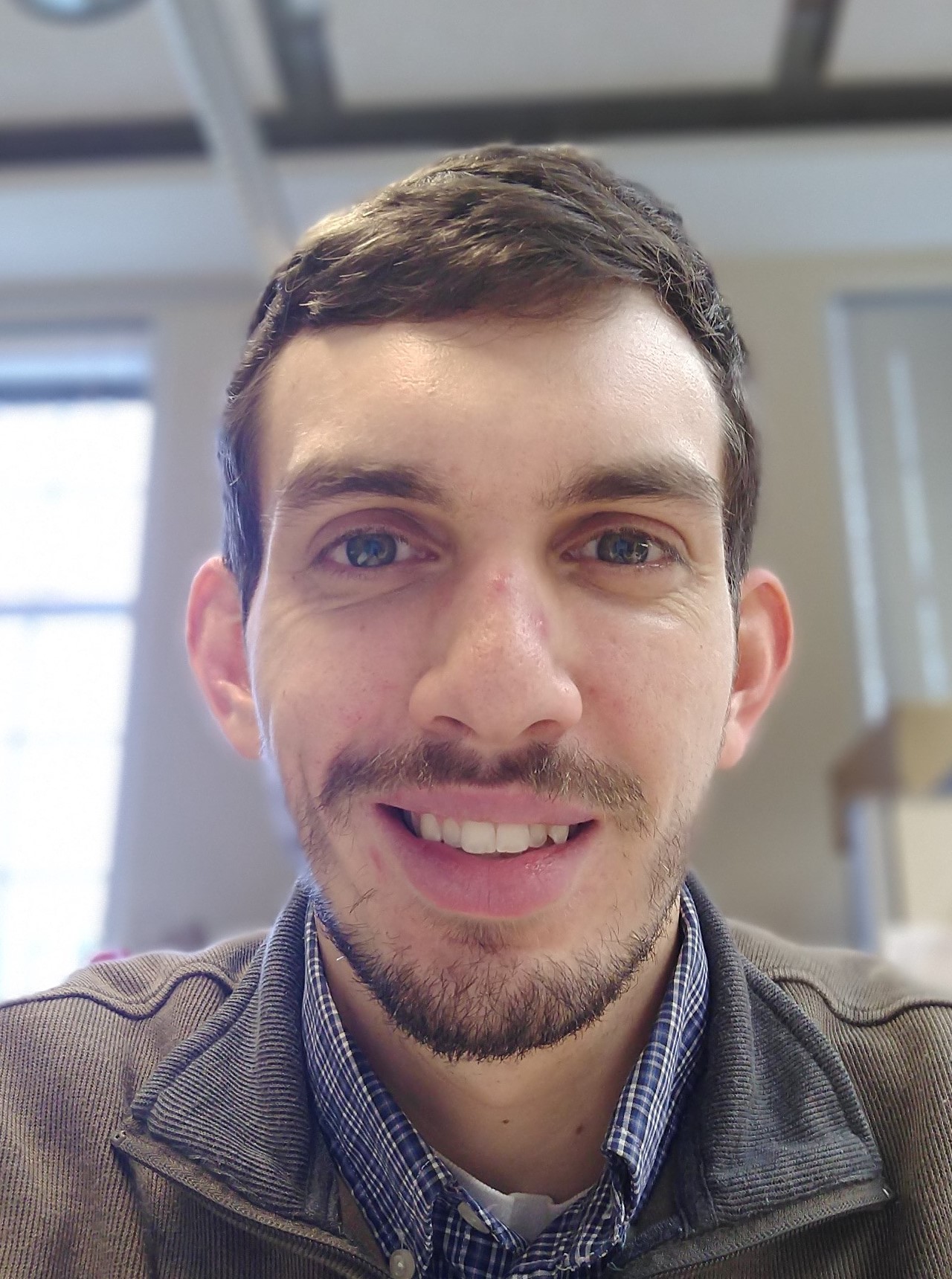}}]{Henry Phalen}
received a B.S. degree in Bioengineering from the University of Pittsburgh in 2018. He is currently a Ph.D. student in Mechanical Engineering at the Johns Hopkins University in Baltimore, MD. His research interests include medical robotics, control systems, systems integration, and medical innovation.
\end{IEEEbiography}

\begin{IEEEbiography}[{\includegraphics[width=1in,height=1.25in,clip,keepaspectratio,angle=0]{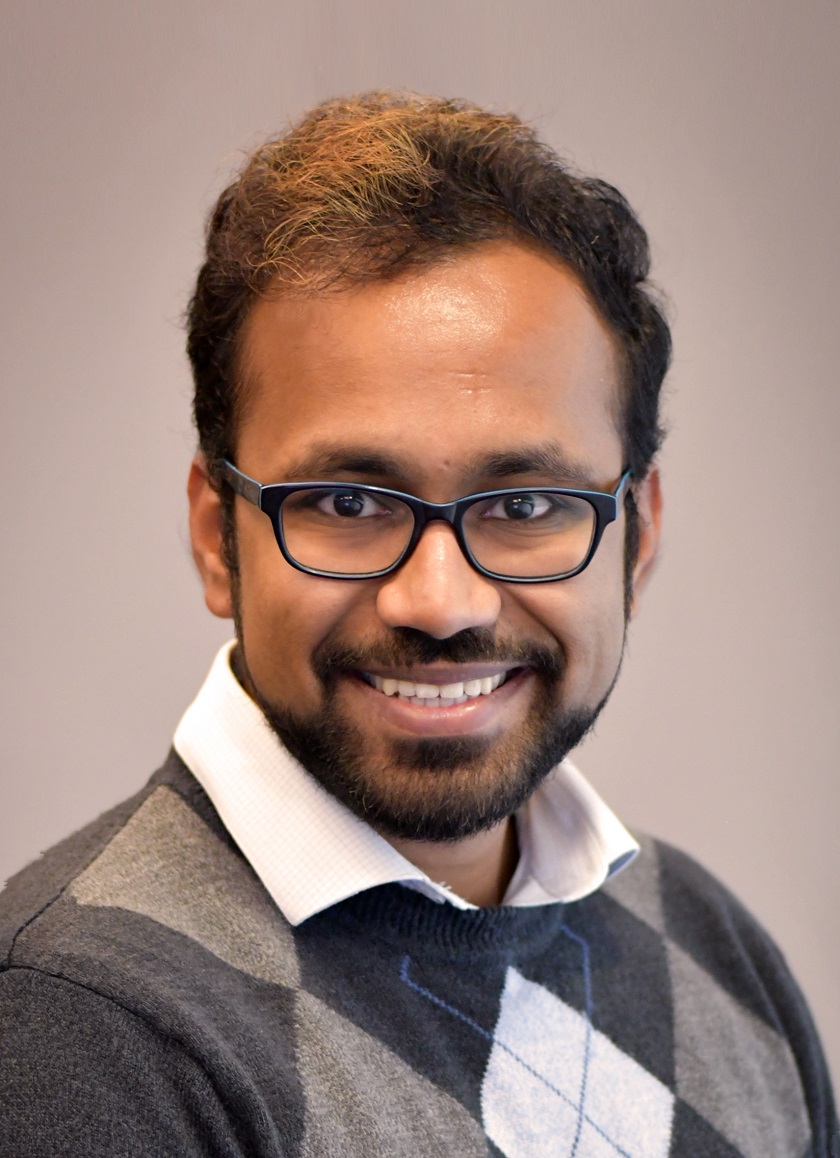}}]{Prasad Vagdargi}
received his B.Tech. (2016) in Mechanical Engineering from Visvesvaraya National Institute of Technology, India and his M.S.E. (2019), specializing in Medical Robotics at Johns Hopkins University, Baltimore, MD. He is currently a PhD student in  Computer Science at I-STAR Lab, Johns Hopkins University. His current research includes surgical computer vision and navigation, image-guided therapy, medical robotics, deep learning, augmented reality and surgical data science.
\end{IEEEbiography}

\begin{IEEEbiography}[{\includegraphics[width=1in,height=1.25in,clip,keepaspectratio,angle=0]{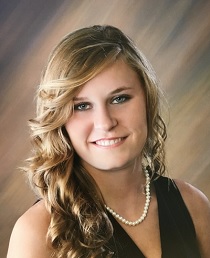}}]{Mariah L. Schrum}
received her B.S. (2018) in biomedical engineering from Johns Hopkins Univ. in Baltimore, Maryland. She is currently working on her Ph.D. in robotics in the CORE Robotics Lab at The Georgia Institute of Technology, Atlanta, GA. Her research interests include utilizing machine learning techniques for medical robotic applications, and she is investigating how we can learn from small heterogenous data sets to improve patient care.
\end{IEEEbiography}

\begin{IEEEbiography}[{\includegraphics[width=1in,height=1.25in,clip,keepaspectratio,angle=0]{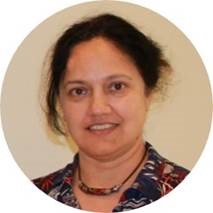}}]{Sumana Chakravarty}
received her PhD in Molecular Biology from the Tata Institute of Fundamental Research in Mumbai, India. She is an expert investigator with formal training in Immunology and over 16 years’ experience in Malaria parasite biology and immunology. She directs immunology research requiring murine and non-human primate models to assess whole sporozoite, recombinant and vectored subunit vaccine platforms against malaria, and combination anthrax, plague and shigella vaccines. She also develops and implements assay methodologies to test the quality of vaccine final product for QC release, for evaluation of humoral responses in Sanaria’s clinical trials, and leads projects in support of automating a key step in Sanaria’s vaccine extraction for Manufacturing.\end{IEEEbiography}

\begin{IEEEbiography}[{\includegraphics[width=1in,height=1.25in,clip,keepaspectratio]{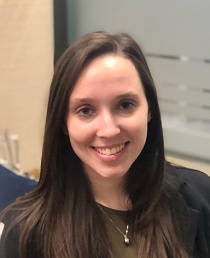}}]{Amanda Canezin}
received a B.S. degree in Mechanical Engineering from the Johns Hopkins University in Baltimore, MD in 2017, where she worked as a Research Assistant in the Laboratory for Computational Sensing \& Robotics (LCSR). Currently, she works at Accenture as a Management Consultant with project experience in digital transformation, business process design \& innovation in the Technology \& Medical Device sectors. 
\end{IEEEbiography}

\begin{IEEEbiography}[{\includegraphics[width=1in,height=1.25in,clip,keepaspectratio]{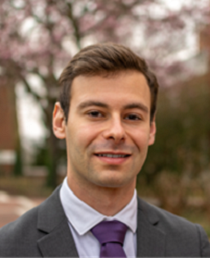}}]{Michael Pozin}
received a B.Sc. degree in Engineering Mechanics and a M.S.E. degree in Biomedical Engineering from the Johns Hopkins University, Baltimore, MD in 2018 and 2019, respectively. He was recently with the Laboratory for Computational Sensing and Robotics, Johns Hopkins University, Baltimore, MD and is currently a Research \& Innovation Engineer with Auris Health Inc., a Johnson \& Johnson family of companies, Redwood City, CA. His research interests include medical robotics, medical devices, and medical instrumentation.   
\end{IEEEbiography}

\begin{IEEEbiography}[{\includegraphics[width=1in,height=1.25in,clip,keepaspectratio]{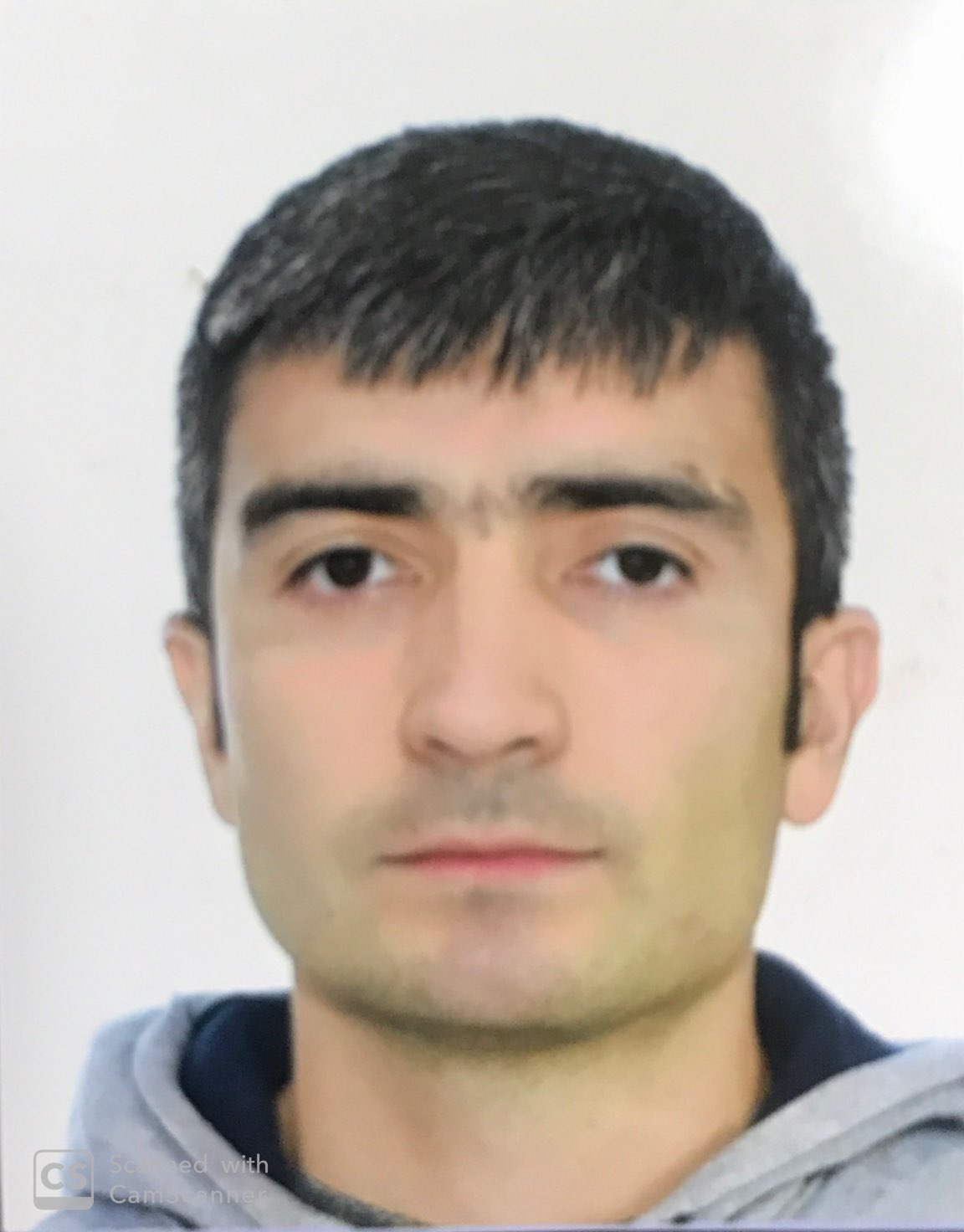}}]{Suat Coemert} earned a B.S. degree in Mechatronics Engineering from Kocaeli University and a M.S. degree in Mechanical Engineering from Koc University in Turkey. He is a research scientist at the Institute of Micro Technology and Medical Device Technology at the Technical University of Munich. His research interests include medical robotics and medical product development. He specializes in design, manufacturing and testing of micromechanical systems developed for medical applications. \end{IEEEbiography}
\begin{IEEEbiography}[{\includegraphics[width=1in,height=1.25in,clip,keepaspectratio]{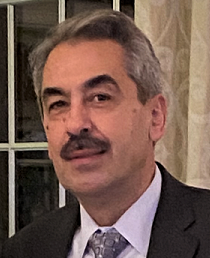}}]{Iulian Iordachita}
(IEEE M’08, S’14) is a core faculty member of the Laboratory for Computational Sensing and Robotics (LCSR) at the Johns Hopkins University, and the director of the Advanced Medical Instrumentation and Robotics Research Laboratory. He received the M.Eng. degree in industrial robotics and the Ph.D. degree in mechanical engineering in 1989 and 1996, respectively, from the University of Craiova. His current research interests include medical/surgical robotics, image-guided surgery, robotics, smart surgical tools, and medical instrumentation.\end{IEEEbiography}

\begin{IEEEbiography}[{\includegraphics[width=1in,height=1.25in,clip,keepaspectratio]{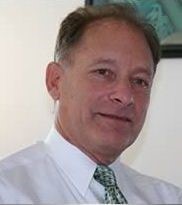}}]{Stephen L. Hoffman}
received his B.A. from University of Pennsylvania, M.D. from Cornell University, Diploma in Tropical Medicine and Hygiene from London School of Hygiene and Tropical Medicine, and did his residency training at U.C. at San Diego. He has over 30 years of experience building and managing large, successful research and development programs. From 1987-2001 he was malaria program director, Naval Medical Research Center. He has held several professorships, chairs or serves on multiple advisory boards, is past president of the American Society of Tropical Medicine and Hygiene, authored over 430 scientific publications, and has numerous patents. He was elected to membership in the National Academy of Medicine in 2004.
\end{IEEEbiography}

\begin{IEEEbiography}[{\includegraphics[width=1in,height=1.25in,clip,keepaspectratio]{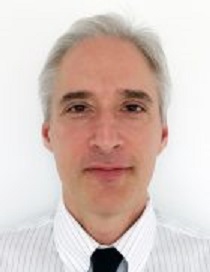}}]{Gregory S. Chirikjian}
(ASME Fellow '08, IEEE Fellow '10) received undergraduate degrees from Johns Hopkins University in 1988, and the Ph.D. degree from the California Institute of Technology, Pasadena, in 1992. He then served on the faculty of the Department of Mechanical Engineering, where he was promoted to the rank of full professor in 2001. From 2004-2007 he served as the JHU ME department chair. Since 2019, he has been serving as the Head of the Department of Mechanical Engineering at the National University of Singapore. His research interests include robotics, applications of group theory in a variety of engineering disciplines, and the mechanics of biological macromolecules.
\end{IEEEbiography}

\begin{IEEEbiography}[{\includegraphics[width=1in,height=1.25in,clip,keepaspectratio]{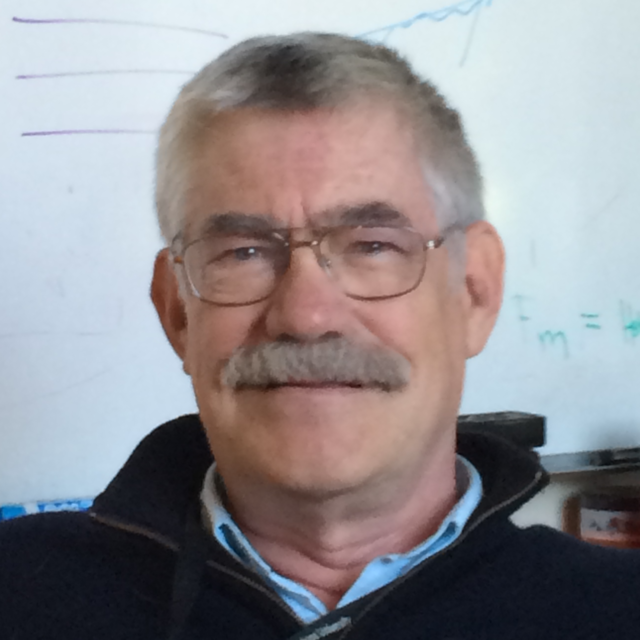}}]{Russell H. Taylor}
has over 30 years’ experience in medical robotics and over 40 in robotics research.  He received his Ph.D. in Computer Science from Stanford in 1976.  After spending 1976 to 1995 as a Research Staff Member and research manager at IBM Research, he moved to Johns Hopkins University in 1995, where he is the John C. Malone Professor of Computer Science with joint appointments in Mechanical Engineering, Radiology, and Surgery and Director of the of the Laboratory for Computational Sensing and Robotics (LCSR).  He is the author of over 450 peer-reviewed journal and conference publications and 83 patents and has received numerous awards and honors including  election to the National Academy of Engineering in 2020.
\end{IEEEbiography}






\end{document}